\documentclass[letterpaper, 10 pt, conference]{ieeeconf}
\IEEEoverridecommandlockouts
\usepackage[acronym,shortcuts]{glossaries}
\usepackage{amsmath,amssymb,amsfonts}
\usepackage{algorithmic}
\usepackage{siunitx}
\usepackage{graphicx}
\usepackage{subcaption}
\usepackage{textcomp}
\usepackage{xcolor}

\usepackage[
backend=biber,
style=ieee,
natbib=true,
maxcitenames=1, 
mincitenames=1,
maxbibnames=3,
]{biblatex}
\usepackage{atbegshi}
\AtBeginShipoutFirst{\raisebox{1cm}{\hspace*{-1in}\makebox[\paperwidth][c]{\footnotesize \textbf{Accepted final version.} IEEE International Symposium on Safety, Security, and Rescue Robotics (SSRR), Galway, Ireland, October 2025}}}
\newacronym{usar}{USAR}{Urban Search and Rescue}
\newacronym{rtf}{RTF}{Robotic Task Force}
\newacronym{drz}{DRZ}{Deutsches Rettungsrobotik Zentrum}
\newacronym{slam}{SLAM}{Simultaneous Localization and Mapping}
\newacronym{sota}{SOTA}{state-of-the-art}
\newacronym{uav}{UAV}{Unmanned Aerial Vehicle}
\newacronym{ugv}{UGV}{Unmanned Ground Vehicle}
\newacronym{lidar}{LiDAR}{Light Detection and Ranging}

\title{\LARGE \bf
Lessons from the Field: A Case Study of Robotic Intervention in an Industrial Emergency
}

\author{
  Jonathan Lichtenfeld${}^{1,*}$,
  Frederik Bark${}^{1,*}$,
  Robert Grafe${}^{2}$,
  Oskar von Stryk${}^{1}$
  \thanks{\raggedright$^{*}$Equal contribution.}
  \thanks{\raggedright$^{1}$The authors are with the SIM Group, Technical University of Darmstadt, Germany. \texttt{\{lichtenfeld,bark,stryk\}@sim.tu-darmstadt.de}}
  \thanks{\raggedright$^{2}$The author is with the German Rescue Robotics Center (DRZ), Germany. \texttt{robert.grafe@rettungsrobotik.de}}
  \thanks{
Research presented in this paper has been supported in parts by the Federal Ministry of Research, Technology and Space (BMFTR) within the DRZ project (grant no. 13N16475), and by the LOEWE initiative (Hesse, Germany) within the emergenCITY center.
 }
}

\begin{document}
\maketitle

\thispagestyle{empty}
\pagestyle{empty}
\begin{abstract}
Incidents in chemical plants can pose a high level of risk and harsh environments for first responders. Contamination and explosion hazards can deny human access to the affected infrastructure, underscoring the need for capable robot systems. 
This field report documents the successful deployment of a robotic task force to neutralize an explosive gas hazard at a chemical plant after a fire incident. 
An \acrfull{ugv} with a custom manipulation tool opened a critical valve under hazardous conditions, averting the threat of a large-scale explosion.
We provide insights into robot deployment and use the mission results to highlight both the importance of rescue robotics and limitations of using research platforms in real emergency deployments, such as communication constraints and the need for enhanced operator-assistance functions.  
\end{abstract}
\section{Introduction}
\begin{figure}
    \centering
    \includegraphics[width=\columnwidth, trim= 0 30 0 0,clip]{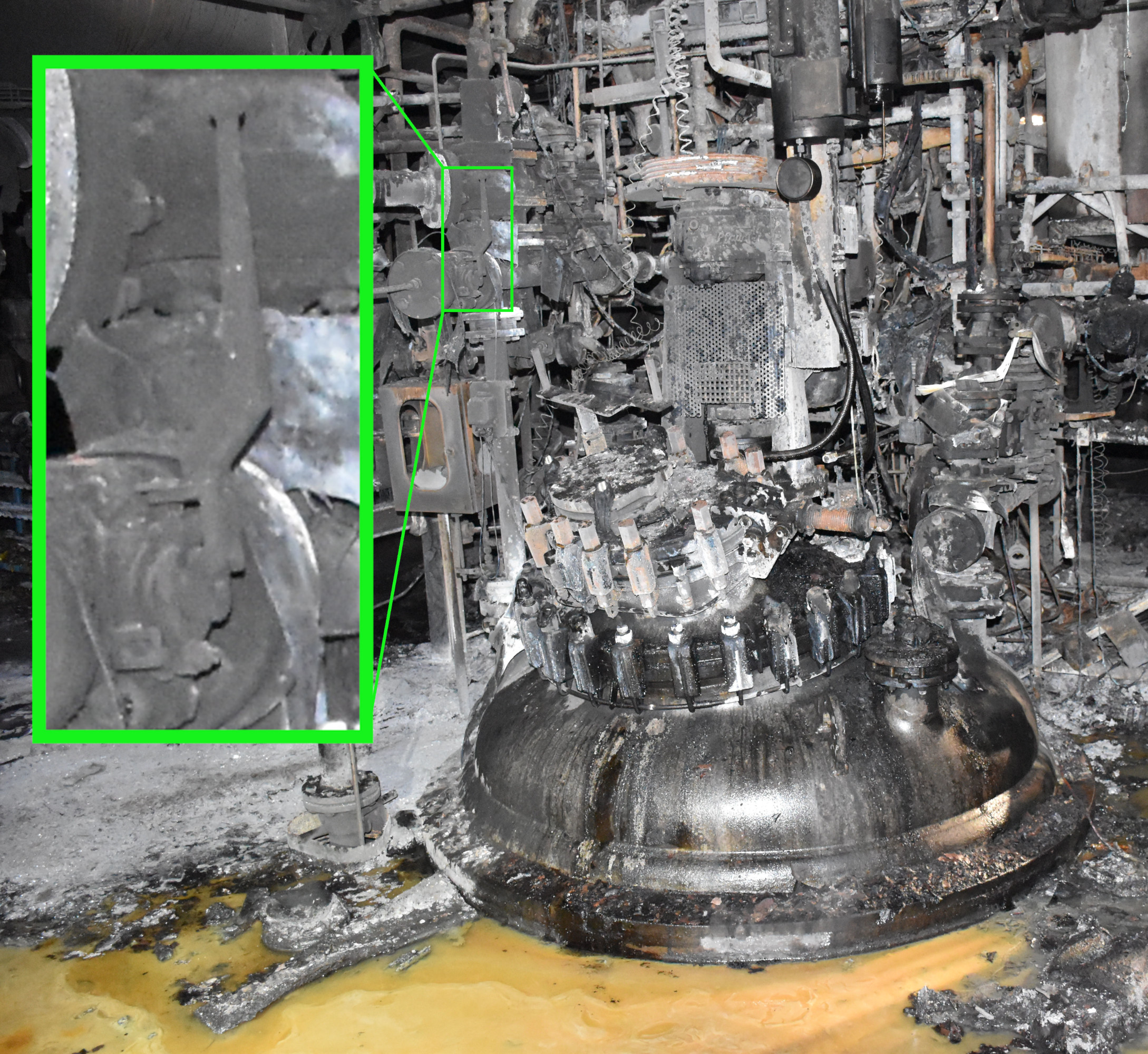}
    \caption{The affected vessel with a close-up of the valve to be opened. The floor is covered with extinguishing water and chemicals. 
    The photo was taken several days before the mission.}
    \label{fig:vessel-firstpage}
\end{figure}

Two weeks after a fire incident in a chemical plant in a major German industrial region
in November 2024,
a highly explosive and shock-sensitive gas was discovered in a damaged reactor. Due to the risk of a large-scale explosion, an immediate response was needed. 
Since the high level of danger excluded human access, a safety radius of approximately 200 meters was established, and the production of the facility was halted.
To resolve the situation and reduce the explosion risk, the plan was to introduce an inerting gas into the reactor, for which a valve next to the reactor had to be opened (Fig. \ref{fig:vessel-firstpage}).
However, the potential formation of friction- or shock-sensitive condensate within the reactor could detonate upon actuation of the valve.

Early in the response phase,
our group
received a request regarding its potential involvement in defusing the situation. 
Initial coordination and situation assessment were followed by a decision to deploy a specialized robotic task force. Assistance was requested through the team's established partner network, resulting in the involvement of the following partner roles:
\begin{itemize}
    \item German Rescue Robotics Center (\acrshort{drz}): Expert consulting, planning, coordination
    \item Regional Emergency Services: \acrfull{uav} operations, logistical support, task force leadership
    \item Local Fire Department and Plant Fire Brigade: On-site logistics and safety management
    \item Technical University of Darmstadt: Operation and adaptation of an \acrshort{ugv}
\end{itemize}

\section{State of the Art}
The importance of robots in \acrfull{usar} and disaster missions has increased over the past decades,
with one of the first widely recognized deployments occurring at the collapsed World Trade Center in 2001 \cite{WorldDisastersReport2017} or inspection and removal operations at the Fukushima-Daiichi Nuclear Power Plant Accident \cite{kawatsumaEmergencyResponseRobots2012a}.
An in-depth historical review is given in \cite{murphyDisasterRobotics2016}.


More recently, research has made advances in autonomous assistance functions that reduce the operator load during a mission \cite{daunRequirementsChallengesAutonomy2023}, such as (supervised) exploration or locomotion and manipulation assistance 
\cite{azayev_2022_autonompus,beck_2009}.
However, a significant gap remains between commercial off-the-shelf systems and research platforms concerning such functionalities \cite{delmerico_2019}.

Our task force is a specialized emergency response unit developed to integrate advanced robotic technologies into civil protection operations. Its mission is to support first responders during hazardous, inaccessible, or complex disaster scenarios—such as chemical incidents, structural collapses, or natural catastrophes—by deploying unmanned ground and aerial systems for safe reconnaissance, manipulation, and data collection. 
Operating under a cooperative model involving national, regional, and municipal authorities, the task force aims to enhance operational efficiency, reduce risks for human responders, and foster the systematic transfer of robotics innovations into real-world emergency management \cite{drz_taskforce_anonymous}.
This report presents a recent successful mission in which our task force actively intervened during an unfolding disaster to prevent further damage.

\section{Mission description}
On the morning of the deployment, the task force assembled at the site 
for a brief meeting with the company's crisis management team. 
This briefing resulted in the finalization of the following mission plan:
\begin{itemize}
    \item The \acrshort{ugv} needs to reach the upper level of the plant.
    \item The damaged reactor and location of the valve must be searched and confirmed.
    \item The valve depicted in Fig. \ref{fig:vessel-firstpage} must be turned by at least 45 degrees so that the inerting gas can be introduced.
    \item Due to limitations in visibility and debris left behind by the fire, a \acrshort{uav} was to provide a second perspective.
\end{itemize}
Additionally, the mission goals were further constrained by 
\begin{itemize}
    \item the required safety distance due to explosion risk, which necessitated reliable remote control of the robot from a safe location
    \item the narrow staircase in the building, making it very difficult for the robot to reach the upper floor by driving
    \item the risk of parts of the upper floor collapsing
    \item the risk of explosion caused by shocks introduced by the \acrshort{ugv} 
    \item the expected distance of the valve at \SI{1.5}{\meter} height and \SI{1.5}{\meter} diameter of the reactor, and the limited reach of the robot arm
    \item the potentially hazardous atmosphere caused by fire, debris, and leaking vessels, containing, for example, hydrochloric acid.
\end{itemize}
While these constraints impeded the deployment and resolution of the mission, they did not hinder the mission's success, and the task force chose deployment under these conditions.

\begin{figure}
\centering
\vspace{4pt}
\includegraphics[width=\columnwidth]{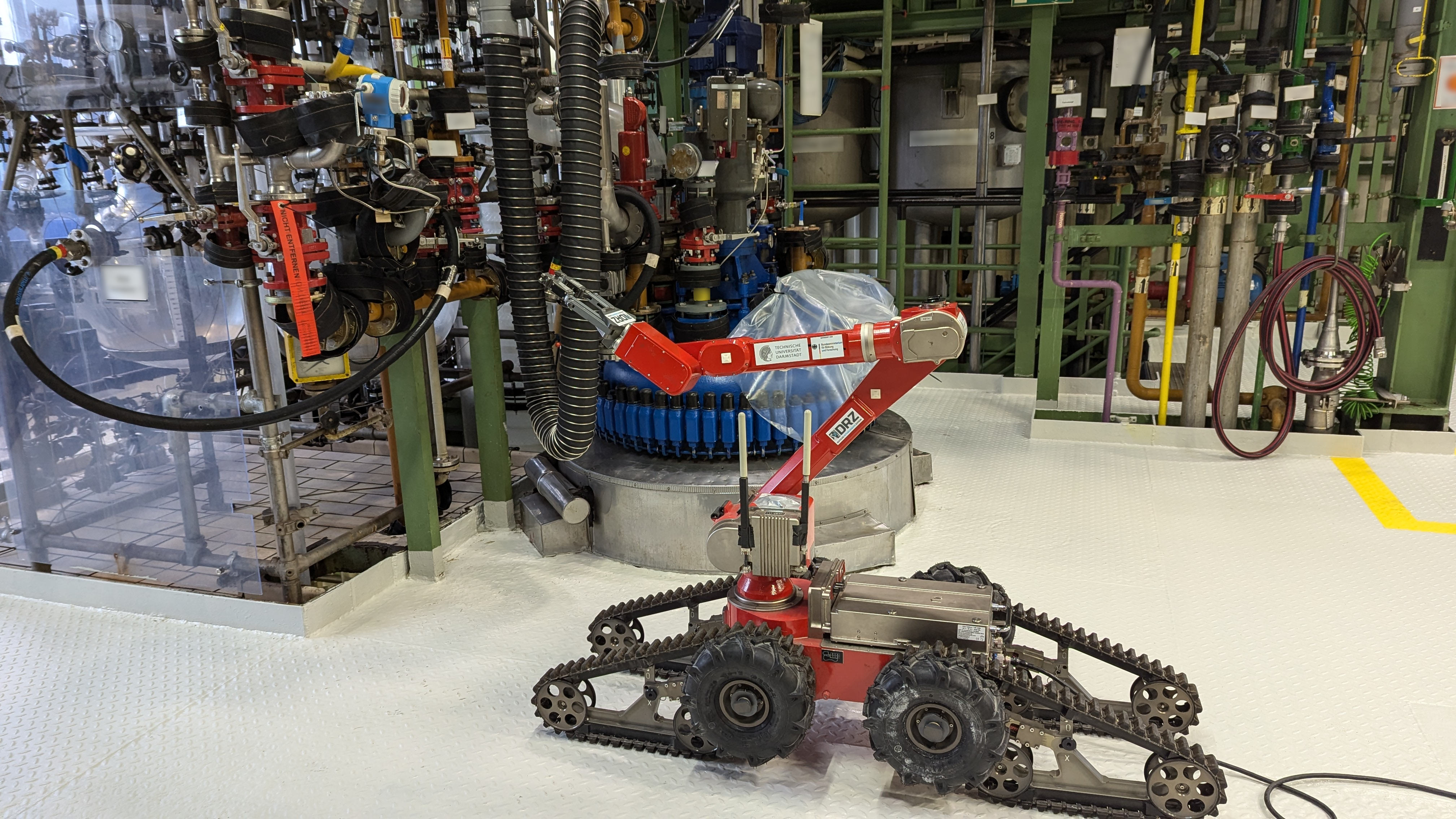}
\caption{Test run with a valve model replicating the real valve that had to be opened during the mission.
}
\label{fig:dry-run}
\end{figure}

\section{Mission preparation}
During mission preparation, the task force and the crisis management split into three groups. 
One group focused on the \acrshort{ugv} deployment, one on the \acrshort{uav} deployment, and one on the further assessment of the scenario concerning suitable locations for safe robot navigation and manipulation.

The selected \acrshort{ugv} for the mission was the Telerob Telemax Hybrid,
chosen due to its compact footprint, which allowed it to be carried upstairs through the stairwell, while also providing the required precision in manipulation. 
Other available robots from first responder units were discussed, but were unsuitable for the mission due to their excessive size and heavier weight.
Additionally, the robot features water and dust protection ratings that enable it to maneuver through muddy debris and residual water resulting from firefighting operations.

\begin{figure}
\centering
\vspace{4pt}
\includegraphics[width=\columnwidth]{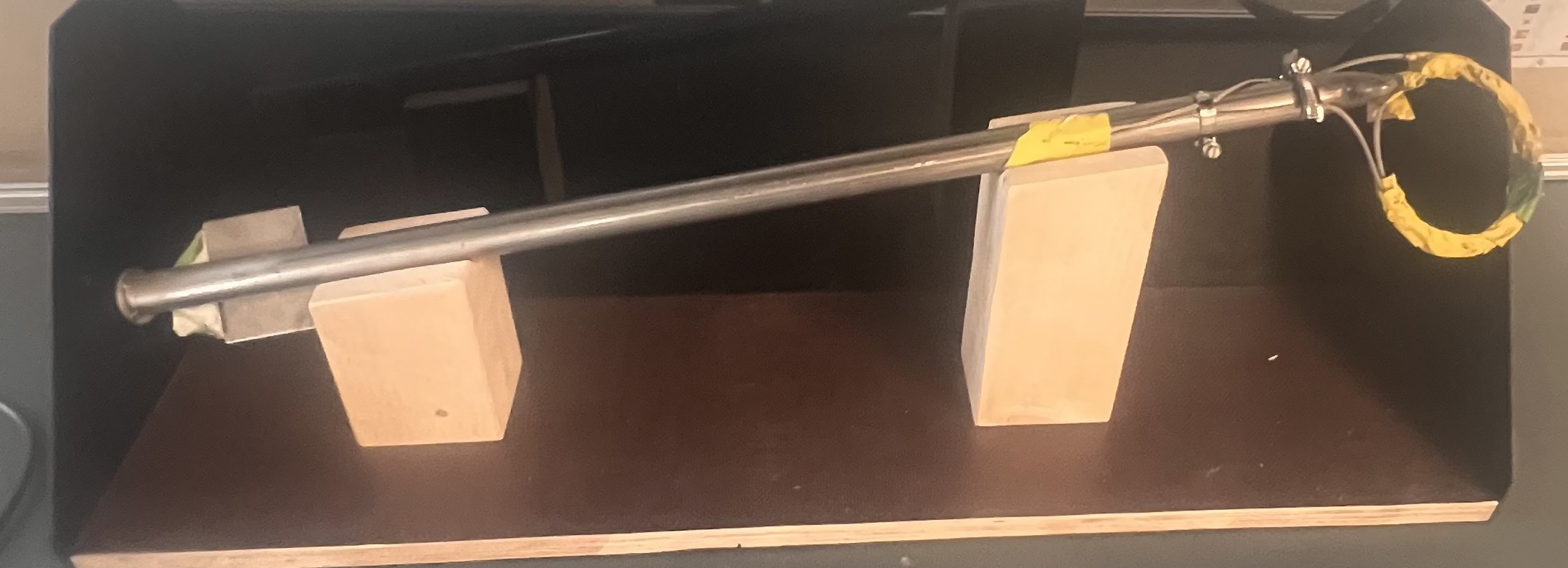}
\caption{
Custom-fabricated manipulator extension with a semi-rigid metal rope loop at the tip, designed to safely operate the valve lever despite limited reach and positioning inaccuracies. The yellow marking enhances tool visibility under low-light conditions.
}
\label{fig:tool}
\end{figure}

The ground robot team conducted dry runs of the manipulation tasks using a similar valve in a different building to verify the robot's ability to reach and pull the valve at the given height of \SI{1.5}{\meter} and expected horizontal distance of \SI{1}{\meter}. 
Fig. \ref{fig:dry-run} shows the robot during the manipulation trial, trying to reach the valve. The robot is placed at a distance from the expected size of the reactor and possible debris near it. 
Although the robot could reach the valve with the use of its flippers, this would entail a certain risk of tipping, which would mean a failure of the mission. 
To resolve this problem, a manipulator extension was custom-fabricated at the company's locksmith shop, enabling the extension of the robot’s last arm segment to the required length.
A loop of metal rope was added at the end to provide a semi-rigid connection. Unlike a fully rigid gripper, which requires precise grasping and exact end-effector trajectories to turn the valve lever, the semi-rigid rope allows for more flexibility in the manipulator’s movements, making it easier to operate the lever despite positioning inaccuracies.
To improve visibility of the tool tip under the dim lighting conditions expected in the mission scenario, it was marked with yellow tape (Fig. \ref{fig:tool}).
The tool was firmly attached to the robot's gripper to reduce the risk of accidental dropping during or after manipulation, as impact shock to the reactor could trigger an explosion.

\begin{figure}
\centering
\includegraphics[width=\columnwidth, trim=0 0 0 35,clip]{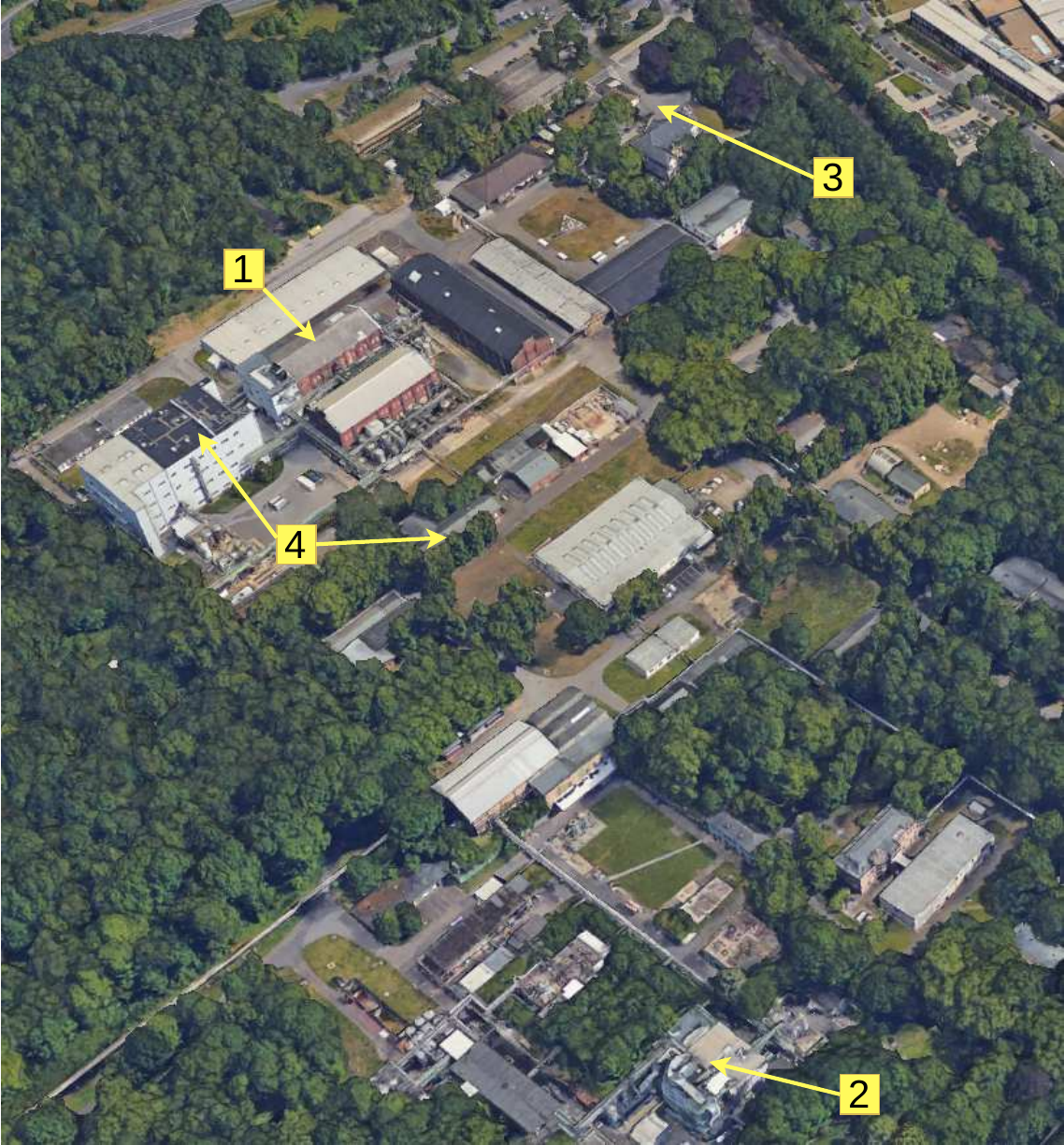}
\caption{Mission site overview; 1 marks the location of the building with the reactor, 2 the position of the ground robot operators, 3 the position of the drone operators, and 4 the positions of the two deployed mesh repeaters.}
\label{fig:scenario-overview}
\end{figure}

The operation distance exceeded \SI{200}{\meter}, requiring the deployment of two radio mesh repeaters on the roofs of surrounding buildings to maintain communication with the \acrshort{ugv}.
Safe operation points at elevated locations outside the safety radius were selected to ensure personnel remained clear of the hazard zone while still providing a good situational overview during the operation.
Due to the limited bandwidth and proprietary interface of the mesh repeaters and corrosive atmosphere inside the building, no additional sensors or research software could be deployed on the ground robot to be used during the mission, limiting its sensors to the cameras integrated by the manufacturer and the control to tele-operation. We will discuss the effects of this restriction in section \ref{sec:discussion}.
A waterproof video camera was mounted for documentation purposes only.
Fig. \ref{fig:scenario-overview} gives an overview of the scenario.
\\

\section{Deployment}

\begin{figure}
\centering
\includegraphics[width=1.0\columnwidth]{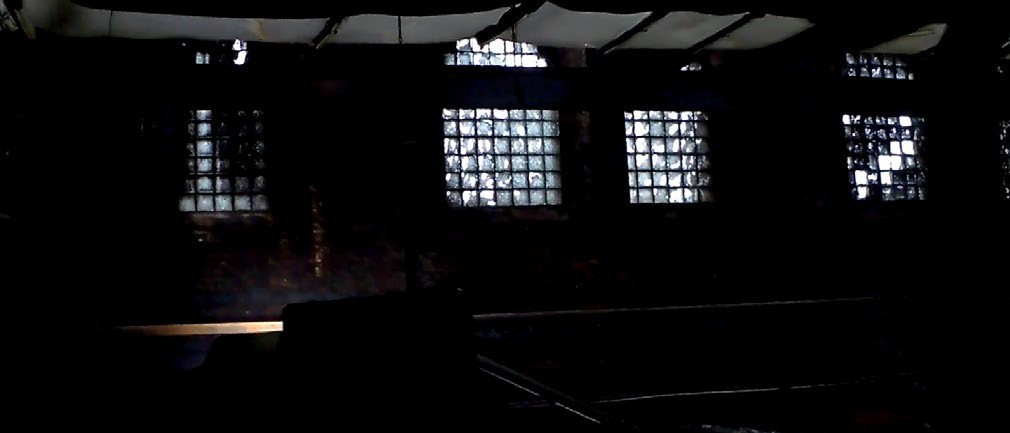}
\caption{\acrshort{ugv} onboard camera view demonstrating challenging visibility conditions inside the facility}
\label{fig:scenario}
\end{figure}

\begin{figure}
\centering
\vspace{4pt}
\begin{subfigure}{0.48\columnwidth}
    \centering
\includegraphics[width=\columnwidth]{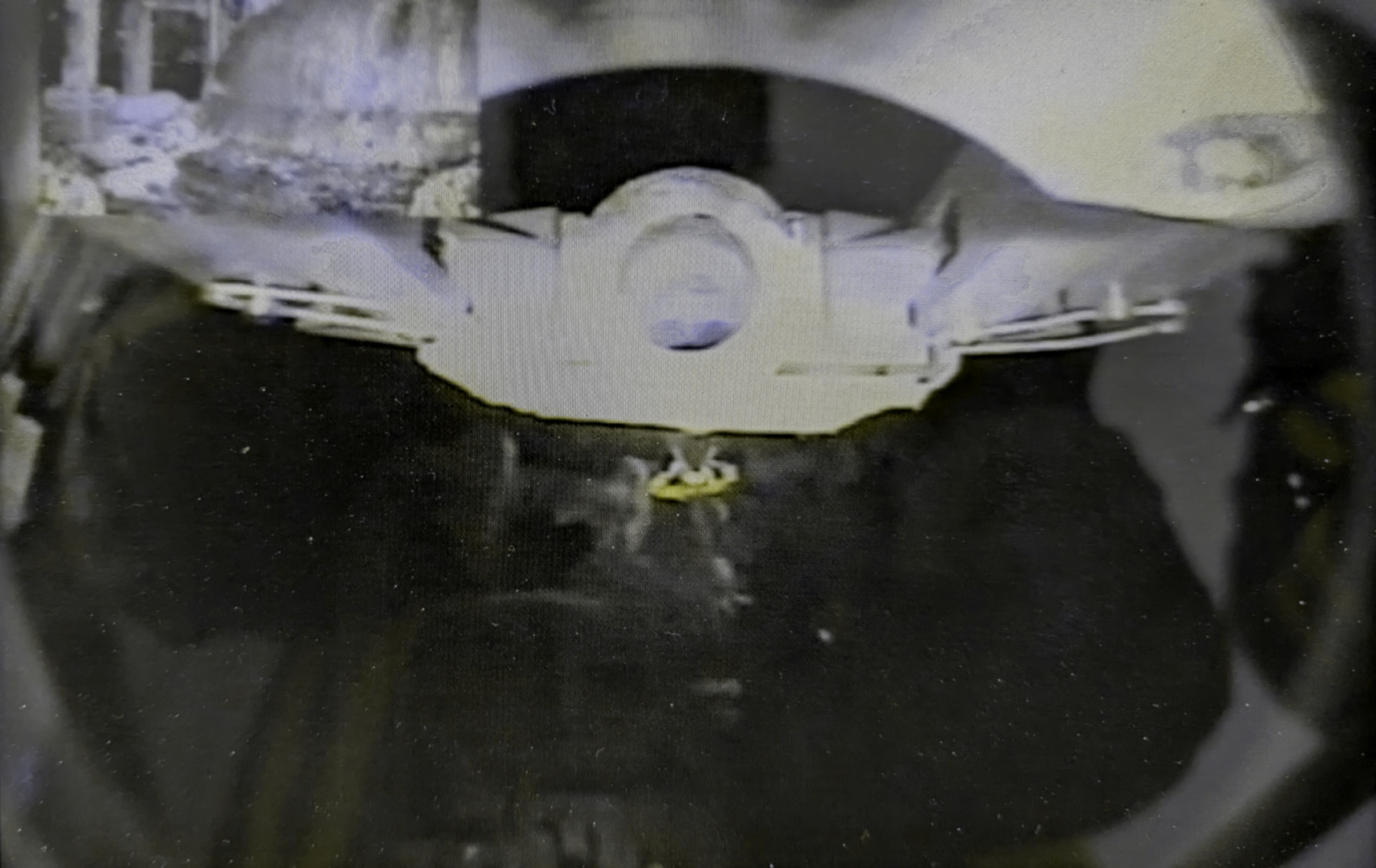}
\end{subfigure}
\begin{subfigure}{0.49\columnwidth}
    \centering
\includegraphics[width=\columnwidth]{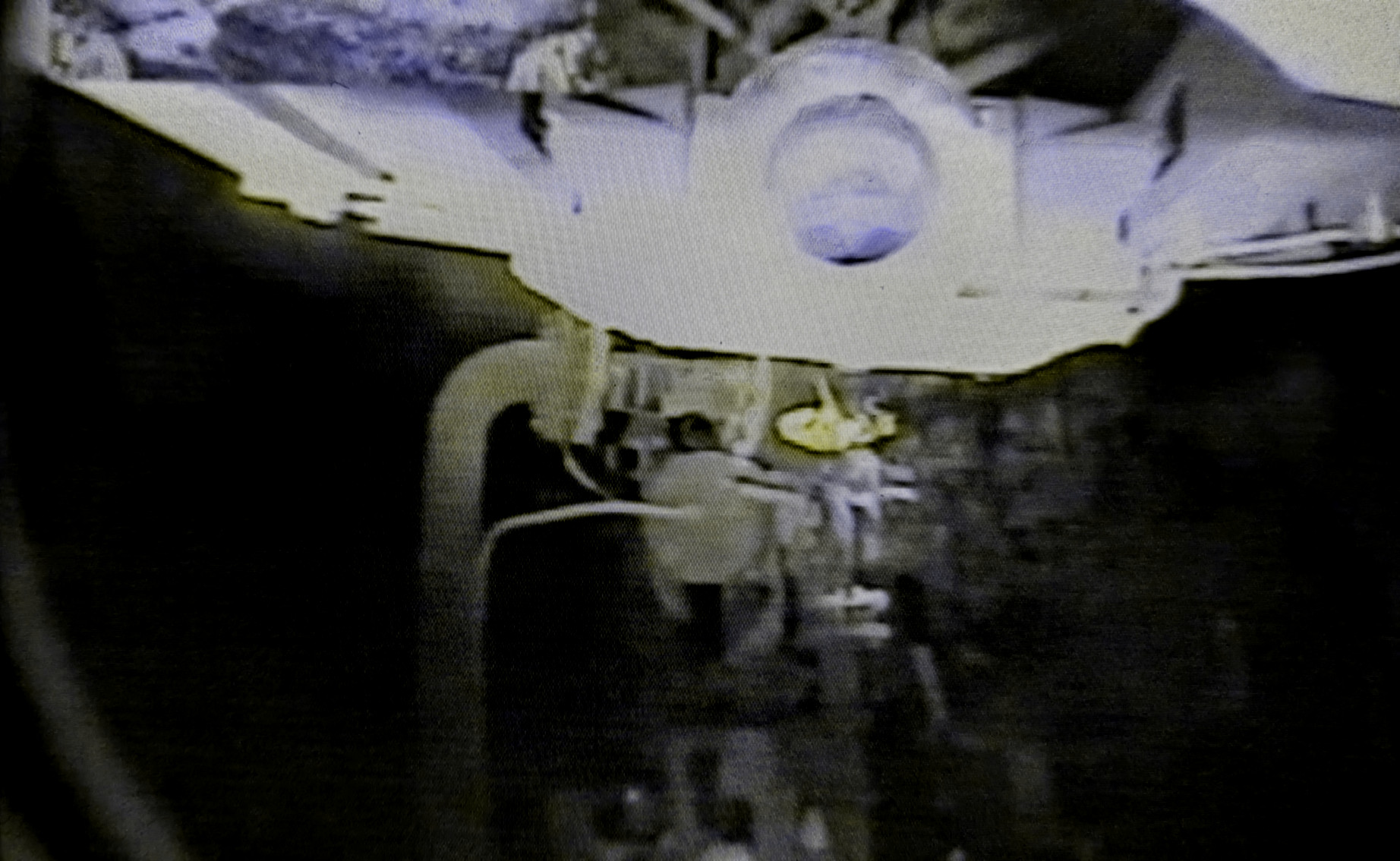}
\end{subfigure}
\caption{The valve during (left) and after (right) manipulation. 
The gripper with the extension tool is visible at the top of the images, including the yellow metal rope at its end.
Image contrast has been enhanced for better visibility.}
\label{fig:manipulation}
\end{figure}

The deployment was executed according to the mission plan, with the UGV operating under challenging environmental conditions.
Fig. \ref{fig:scenario} shows a picture taken by the external video camera during the exploration of the building.
Due to the late autumn conditions, there was little light in the building.
The operators were able to navigate the robot around small pieces of debris and some metal pieces hanging from the ceiling with the help of lights near its front, rear, and gripper, as well as an additional flashlight on its arm. 
A suitable position for manipulation in front of the valve was reached after approximately \SI{10}{\minute}.
Small amounts of debris in front of the reactor as well as nearby pipes hindered the manipulation, but with the help of the extension tool, the valve could be opened as shown in Fig. \ref{fig:manipulation}. 

\begin{figure}
\centering
\vspace{4pt}
\includegraphics[width=\columnwidth]{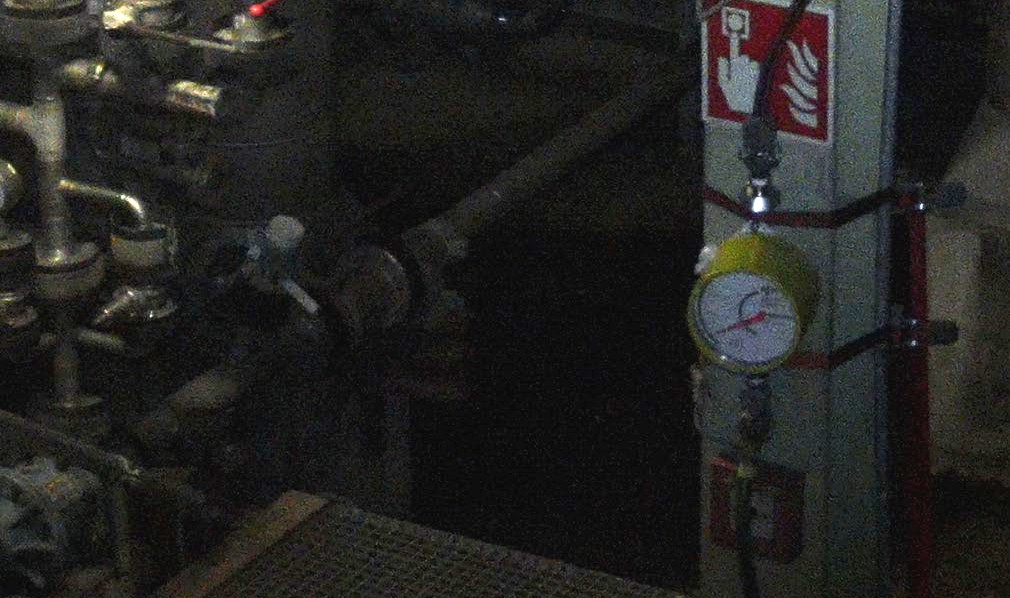}
\caption{Image captured by the drone confirming the flow of the inerting gas.}
\label{fig:drone}
\end{figure}

Besides the UGV deployment, members of 
the local fire brigade also utilized two drones to assist in reconnaissance. An indoor drone was initially intended to provide visual support for the teleoperation of the UGV. Due to insufficient radio penetration within the facility, the interior flight was aborted. Instead, a second drone was repurposed for external observation of a flowmeter to confirm successful inert gas transfer following the valve activation (Fig. \ref{fig:drone}).

During the deployment, a member of the on-site crisis team assisted the \acrshort{ugv} operators by 
guiding them through the scenario, confirming the identified reactor, and helping determine the correct valve to open. This support decreased the time spent on site and contributed to the mission's success.

Post-operation decontamination was necessary to remove hazardous contaminants and burn residue from the robot. The corrosive environment damaged seals and tracks, requiring replacement of affected components.

\begin{figure}
\centering
\vspace{4pt}
\includegraphics[width=\columnwidth]{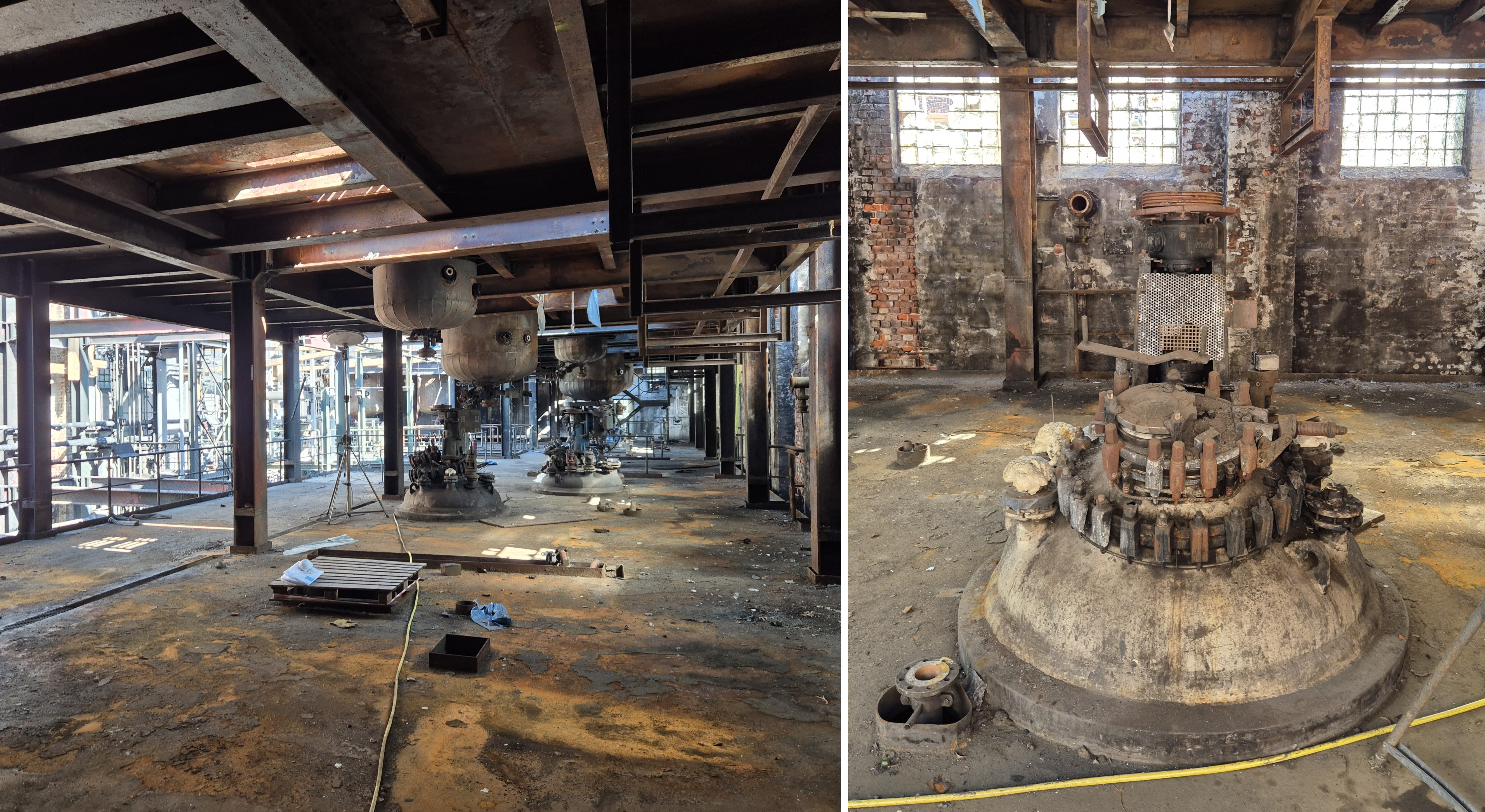}
\caption{Post-incident view of the facility. \textbf{Left:} Overview of the building, showing the floor where the accident occurred. \textbf{Right:} Close-up of the vessel involved in the incident (see Fig. \ref{fig:vessel-firstpage}). Burn and corrosion damage remain clearly visible.}
\label{fig:aftermath:overview}
\end{figure}

\begin{figure}
\centering
\vspace{4pt}
\includegraphics[width=\columnwidth]{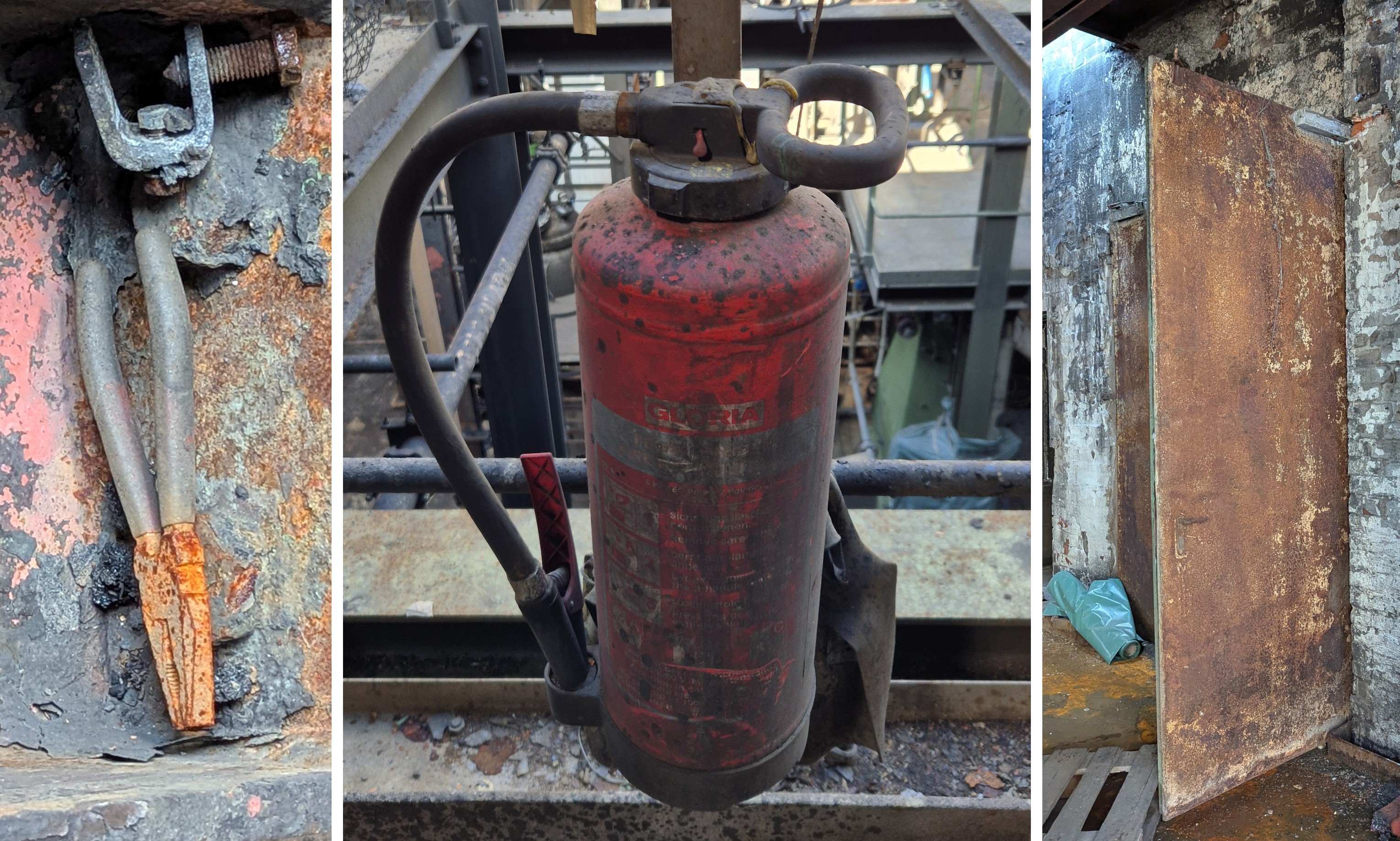}
\caption{Examples of objects severely damaged during the incident. \textbf{Left:} Pliers. \textbf{Middle:} Fire extinguisher. \textbf{Right:} Door used as the entry point to the building during the mission.}
\label{fig:aftermath:objects}
\end{figure}

In order to further document the mission environment, an additional data collection was carried out approximately one year after the incident.
Detailed images (Fig. \ref{fig:aftermath:overview}, \ref{fig:aftermath:objects}) and a 3D point cloud of the building (Fig. \ref{fig:pointcloud_topwdown}) were recorded to capture the site conditions after the intervention.
In the meantime, parts of the building structure had been reinforced to prevent collapse, and several vessels and tubes had been removed. 
Nevertheless, the overall environment remained burned, rusted, and heavily degraded.
These data complement the original deployment observations and provide a valuable basis for subsequent analysis and dissemination.

\section{Discussions}
\label{sec:discussion}
For research purposes, a module box with custom sensors had been developed and integrated on the robot prior to this event to circumvent restrictions of the proprietary robot interfaces.
This box contains a variety of sensors such as RGB-D cameras, a \acrfull{lidar} sensor, thermal imaging cameras, and mobile computers.
In addition, it offers a fully ROS2-integrated software stack to provide the ground robot operator with state-of-research software for urban search and exploration.
However, this module box relies on Wi-Fi communication to connect the robot and the operator station. Setting up a Wi-Fi link over long distances was not feasible within the available time. Only the manufacturer’s internal radio system could cover this range using mesh nodes.
As a result, communication was limited to the proprietary radio module, without onboard software or additional sensors. Consequently, the available assistance functions could not be used.
Furthermore, the research module was not hardened for deployment in environments with contamination risks from corrosive atmospheres, rain, or firefighting water, and was therefore removed.
The operator station provided by the robot manufacturer offers video streams from three cameras (front, rear, and gripper) as well as a visualization of the current joint configuration.

In the following, we discuss issues encountered during the mission and how they could have been addressed using current research approaches.

\subsection{Using research platforms in deployment}
Research platforms are often built without hardened housing. 
This enables faster iterations and adjustments to the hardware as less sealing material needs to be reapplied, and the time required for development is reduced.
As a result, dust, liquids, and gases can potentially damage expensive components and render them unusable. 
Similarly, research software often falls short of the reliability and quality of commercial software.
It aims to advance the state of research, but cannot guarantee 100\% reliable operation.
This is of crucial importance in risky applications where software failures could impair functionality and lead to damage to the robot, mission site, or bystanders.

\subsection{State-of-research support systems}
Direct teleoperation in robotic deployment is time-consuming, prone to human error, and requires well-trained operators \cite{murphyDisasterRobotics2016}.
Following the definition by Daun et al.~\cite{daunRequirementsChallengesAutonomy2023}, \textit{perception assistance} and \textit{control assistance} can reduce the operator load and improve performance.
Beyond assistance functions, robust \textit{communication} and systematic \textit{mission documentation} play a crucial role in ensuring overall mission success and traceability.

\subsubsection{Perception Assistance}
Tele-operated missions demand a high degree of situational awareness on the part of the operator to ensure accurate navigation, manipulation, and decision-making.
Perceptional situation awareness and support of operators can be provided with methods such as:
\begin{itemize}
    \item \textbf{\acrfull{slam}} is a crucial capability for navigation and route planning. State-of-the-art, mostly lidar- \cite{koideGLIM3DRangeInertial2024a} or camera-based \cite{mur-artalORBSLAMVersatileAccurate2015} algorithms create detailed 3D maps of the surroundings and localize the robot down to the centimetre. \acrshort{slam} systems are available on some modern robots, but often not adapted to the harsh conditions encountered in search and rescue missions.
    As an illustration, Fig.~\ref{fig:pointcloud_topwdown} shows a point cloud map of the damaged building, which exemplifies the type of situational information that could be available to an operator during a mission.

    \begin{figure}
    \centering
    \vspace{4pt}
    \includegraphics[width=\columnwidth]{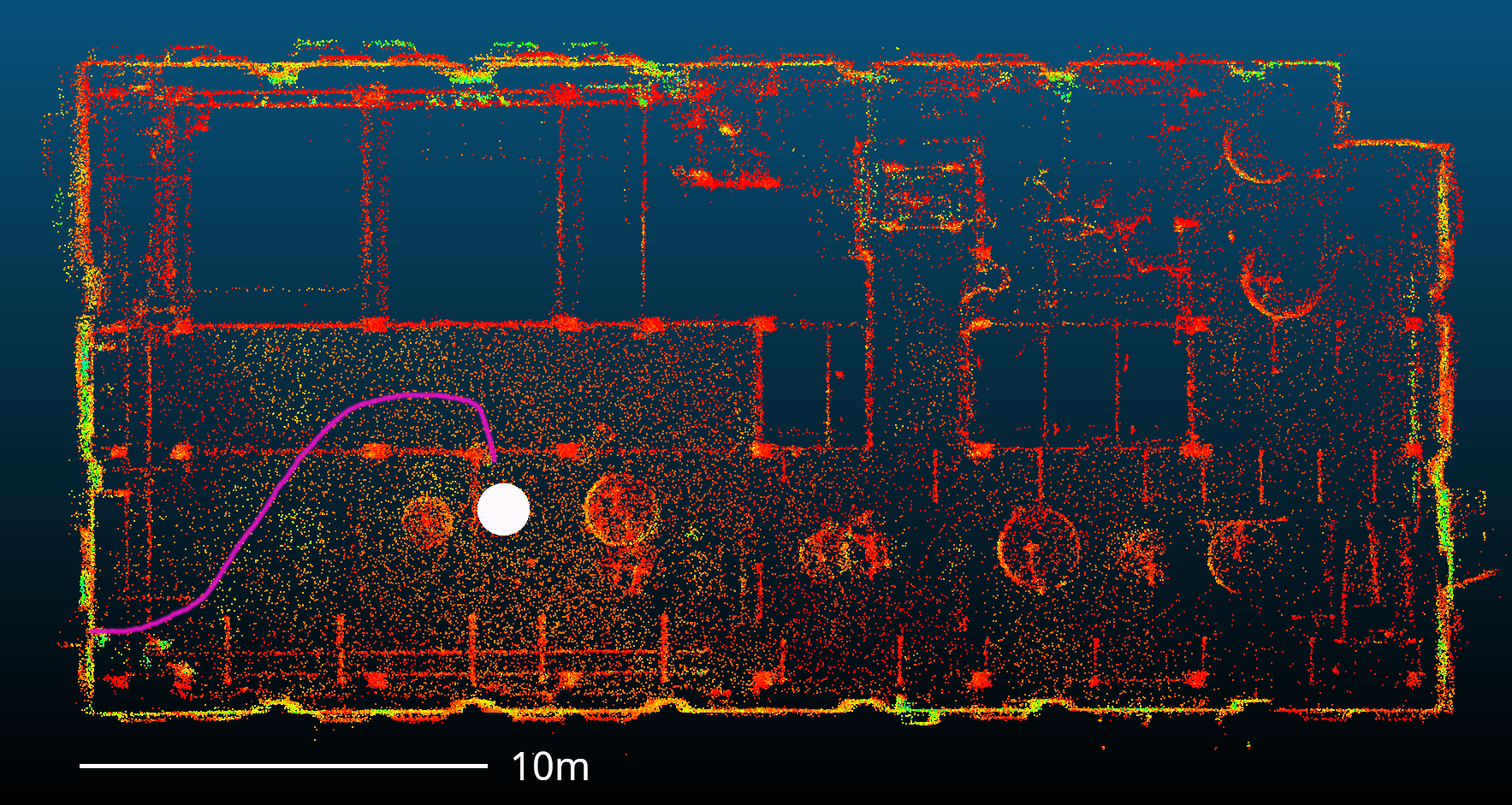}
    \caption{Point cloud reconstruction of the damaged building acquired with a handheld LiDAR sensor during the follow-up inspection. The roof was removed for clarity, showing the second floor. Pink indicates the approximate robot trajectory, while white marks the location of the damaged vessel.}
    \label{fig:pointcloud_topwdown}
    \end{figure}
    \item \textbf{Multi-modal visual perception} provides operators with intuitive and complementary information, such as depth and thermal data, which are especially valuable in low-visibility conditions caused by darkness, smoke, or dust. State-of-the-art perception systems that combine visual, depth, and thermal modalities could have significantly improved hazard awareness and object identification \cite{ulloa_2024_thermal}, for instance by distinguishing between structural debris and manipulable objects. Looking ahead, we are considering collecting and analyzing images of objects damaged or destroyed by fire from this (see Fig. \ref{fig:aftermath:objects}) and similar incidents to explore their potential use in evaluating standard object detection methods. Since such burned or deformed objects are rarely represented in existing training datasets, detection performance is likely to degrade. Such an analysis could provide valuable insights into how perception algorithms behave in post-disaster environments and guide the development of more robust hazard detection systems.
    In our mission, perception was limited to the robot’s built-in cameras, without access to raw sensor data for advanced processing.
    
    \item Heterogeneous \textbf{multi-agent teams} can complement each other's strengths \cite{drewMultiAgentSystemsSearch2021}. 
    \acrshort{uav}s are well-suited for rapid exploration of large areas, whereas \acrshort{ugv}s provide higher payload capacity and advanced manipulation capabilities. 
    Typical collaborations include \acrshort{uav}s mapping an area and subsequently guiding \acrshort{ugv}s through it \cite{muegglerAerialguidedNavigationGround2014}, or providing alternative viewpoints by positioning near the \acrshort{ugv} and transmitting live video to assist with navigation or manipulation. 
    In our mission, an additional viewing angle was planned to support the complex manipulation task, but this could not be realized due to insufficient wireless communication for the \acrshort{uav}.

\end{itemize}

\subsubsection{Control Assistance}
To reduce the operator’s burden, various support functions have been developed that enhance navigation and stability through automated or semi-automated control mechanisms.
\begin{itemize}
    \item \textbf{Locomotion assistance} includes capabilities such as traversability estimation, collision avoidance, and stability guidance. 
    These functions support safer and more efficient navigation, especially in poor lighting or when debris is not visible in the camera images. 
    Traversability estimation \cite{sevastopoulosSurveyTraversabilityEstimation2022} and collision avoidance with the environment \cite{alajlanMultisensorBasedCollision2015} help avoid hazards, while stability guidance \cite{oehlerAccuratePosePrediction2023} reduces the risk of tip-over. 
    
    Since these features were not available, the operators had to proceed with extreme caution.
    \item \textbf{Navigation assistance} can further reduce the operator’s workload by enabling automatic travel to user-defined waypoints or even autonomous frontier-based exploration \cite{nahavandi_2025_comprehensive}. 
    Such capabilities would have allowed the operators to focus more on situational awareness and environmental monitoring. 
    Since the position of the damaged reactor was already known beforehand, waypoint navigation would have been particularly useful. 
    Instead, the robot had to be manually teleoperated through difficult terrain and poor visibility, which resulted in a \SI{10}{\minute} approach to the valve despite the short distance (see Fig.~\ref{fig:pointcloud_topwdown}). 
    During this process, the operators were directly supported by a member of the plant’s crisis team, who identified the reactor in the robot’s camera images.
    Full autonomous navigation would build upon the above-mentioned capabilities, including \acrshort{slam}, stability/traversability estimation, and locomotion assistance.
    
\end{itemize}
\subsubsection{Communication}
One of the strongest limitations during the mission was the lack of a reliable ad-hoc Wi-Fi network for sensor data transfer, underscoring the need for robust communication systems. 
In disaster scenarios, communication is often hampered by damaged infrastructure, metallic structures, and environmental conditions such as smoke, dust, or water. 
Without a stable link, valuable sensor data from external modules cannot be transmitted to the operator, effectively restricting situational awareness to the proprietary interfaces provided by the base robot.

State-of-the-art approaches to improve communication in these settings include the use of relay nodes or mesh networks, where multiple ground or aerial robots extend coverage dynamically. 
Alternative technologies, such as private 4G/5G networks or directional antennas, can also provide higher bandwidth and more reliable connections in complex environments. 
In practice, the deployment of such systems requires rapid setup, interoperability with existing hardware, and resilience against harsh environmental conditions. 

For the reported mission, setting up a Wi-Fi network on the fly to cover the safety distance to the building was not feasible, partly because the building could not be entered, and brick walls reinforced with steel structures severely attenuated the signal.
As a result, communication relied solely on the robot’s proprietary low-bandwidth radio, which provided only basic video and control streams. 
This restriction limited the integration of additional perspectives, such as UAV support, and underscored the need for portable, rapidly deployable communication infrastructures that can ensure reliable, long-range coverage in future deployments.

\subsubsection{Mission Documentation}
Post-deployment data evaluation is an integral part of rescue missions. 
As the robot was used without our sensor module, only a low-quality dash cam was available for video recording.
An alternative approach to address contamination and hardware damage risks would be to develop lower-cost "disposable" modules with essential sensors for data recording and operation assistance.
In our case, this would include a \acrshort{lidar} for 3D reconstruction and cameras for inspection.
This information could help to further assess the situation in the aftermath, even if no data transfer was possible during deployment.

\section{Conclusions}
This field report describes the successful deployment of a \acrshort{ugv} in an industrial emergency, which likely helped prevent significant damage. 
We highlight the limitations encountered when applying state-of-the-art techniques in real-world operations and discuss how these approaches could enhance performance and safety if fully integrated. 
Our observations emphasize the continued importance of robust hardware, reliable communication, and comprehensive mission documentation to support both human operators and autonomous systems in challenging environments.


\addtolength{\textheight}{-6cm}   

\renewcommand*{\bibfont}{\footnotesize}
\printbibliography

\end{document}